\documentclass[10pt,twocolumn,letterpaper]{article}

\usepackage[pagenumbers]{cvpr} % To force page numbers, e.g. for an arXiv version

\usepackage{graphicx}
\usepackage{amsmath}
\usepackage{amssymb}
\usepackage{booktabs}

\usepackage{multirow}

\usepackage{xcolor}

\newcommand{\best}[1]{\textbf{#1}}

\definecolor{Gray}{gray}{0.95}
\usepackage{colortbl}
\newcolumntype{a}{>{\columncolor{Gray}}c}
\newcommand{\mc}[2]{\multicolumn{#1}{c}{#2}}

\newcommand{\floor}[1]{\lfloor #1 \rfloor}
\usepackage{pifont}   %cross and tick mark
\newcommand{\cmark}{\ding{51}}%
\newcommand{\xmark}{\ding{55}}%

\usepackage{pifont}   %cross and tick mark
\usepackage[pagebackref,breaklinks,colorlinks]{hyperref}

\usepackage[capitalize]{cleveref}
\crefname{section}{Sec.}{Secs.}
\Crefname{section}{Section}{Sections}
\Crefname{table}{Table}{Tables}
\crefname{table}{Tab.}{Tabs.}

\newcommand*{\affaddr}[1]{#1} 
\newcommand*{\affmark}[1][*]{\textsuperscript{#1}}
\newcommand*{\email}[1]{\texttt{#1}}
\usepackage[symbol]{footmisc}

\def\cvprPaperID{*****} % *** Enter the CVPR Paper ID here
\def\confName{xxxx}
\def\confYear{2023}

\begin{document}

\title{ToonCrafter: Generative Cartoon Interpolation}

\author{%
Jinbo Xing\affmark[1]~~~~~~~~~~~ 
Hanyuan Liu\affmark[2]~~~~~~~~~~~
Menghan Xia\affmark[3,]$^*$~~~~~~~~~~~
Yong Zhang\affmark[3]\\
Xintao Wang\affmark[3]~~~~~~~~~~~
Ying Shan\affmark[3]~~~~~~~~~~~
Tien-Tsin Wong\affmark[1,]$^*$\\
\affaddr{\affmark[1]CUHK~~~~~~~~~~}
\affaddr{\affmark[2]CityU~~~~~~~~~~}
\affaddr{\affmark[3]Tencent AI Lab~~~~~~~~~~}\\
% \small{\email{\{jbxing,yczhang21,ttwong\}@cse.cuhk.edu.hk ~~~~ \{menghanxyz,vinthony,arphid\}@gmail.com}}
\small\email{$^*$Corresponding Author. Project page: \url{https://doubiiu.github.io/projects/ToonCrafter}}
}

\twocolumn[{
\maketitle

\begin{center}
    \captionsetup{type=figure}
    \includegraphics[width=1\textwidth]{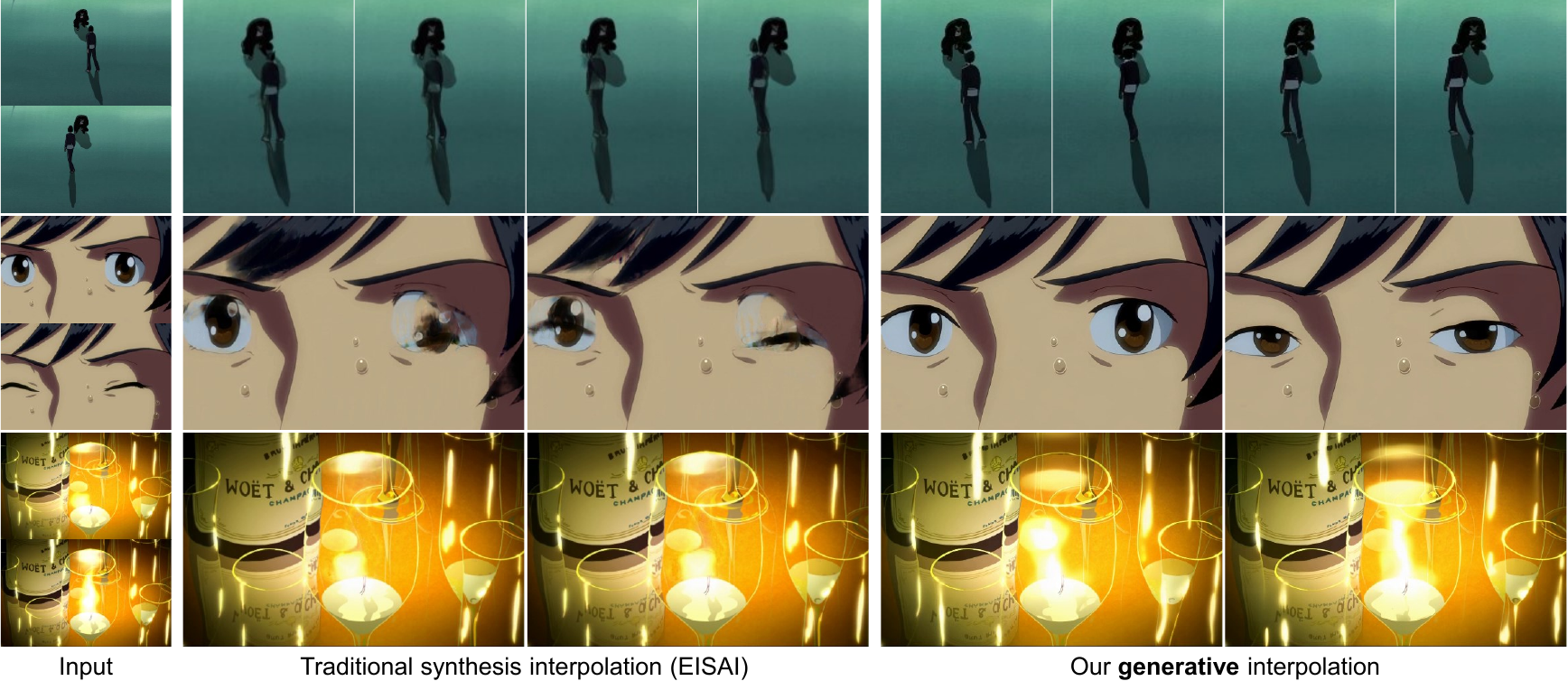}
    \captionof{figure}{
    }
    \label{fig:teaser}
\end{center}
}]
%%%%%%%%% ABSTRACT
\begin{abstract}
We introduce ToonCrafter, a novel approach that transcends traditional correspondence-based cartoon video interpolation, paving the way for generative interpolation. Traditional methods, that implicitly assume linear motion and the absence of complicated phenomena like dis-occlusion, often struggle with the exaggerated non-linear and large motions with occlusion commonly found in cartoons, resulting in implausible or even failed interpolation results. To overcome these limitations, we explore the potential of adapting live-action video priors to better suit cartoon interpolation within a generative framework.
ToonCrafter effectively addresses the challenges faced when applying live-action video motion priors to generative cartoon interpolation. First, we design a toon rectification learning strategy that seamlessly adapts live-action video priors to the cartoon domain, resolving the domain gap and content leakage issues. Next, we introduce a dual-reference-based 3D decoder to compensate for lost details due to the highly compressed latent prior spaces, ensuring the preservation of fine details in interpolation results. Finally, we design a flexible sketch encoder that empowers users with interactive control over the interpolation results.
Experimental results demonstrate that our proposed method not only produces visually convincing and more natural dynamics, but also effectively handles dis-occlusion. The comparative evaluation demonstrates the notable superiority of our approach over existing competitors.
Code will be available upon publication.
\end{abstract}

%%%%%%%%% BODY TEXT
\section{Introduction}
\label{sec:introduction}

While cartoon animation is a popular medium and art form,  its production remains labor-intensive due to its frame-by-frame manual drawing nature.
Recent live-action video frame interpolation methods~\cite{reda2022film,bao2019depth} can automate the production to certain degrees, but far from satisfactory for cartoon frame interpolation. The differences between cartoon animation and live-action video lie in two aspects, the frame ``sparsity'' and the texture richness. While live-action video frames can be densely acquired by camera, cartoon frames are temporally sparse (hence, large motion) due to the high drawing cost. Such cost also leads to higher chance of textureless color regions in cartoon than in live-action video. Both characteristics  make  cartoon frame interpolation much challenging.

Over the past few years, cartoon video interpolation~\cite{chen2022eisai,siyao2021anime} has been significantly advanced by deep neural networks.
However, these developments have primarily focused on linear interpolation by assuming the underlying motion is straightforward.
Their general idea is first to identify correspondences between two frames, e.g. optical flow~\cite{teed2020raft}, and then  linearly interpolate the frames. 
Figure~\ref{fig:teaser}(top) shows an example featuring a walking person, apparently linear interpolation can only generate a ``drifting'' person, instead of a correct walking sequence. 
Things get even more complicated in a dis-occlusion example in Figure~\ref{fig:teaser}(middle).
In other words, linear motion assumption is largely insufficient for interpolating obvious motions observed in our daily life. 
In this paper, we point out the importance of {\bf generative} cartoon interpolation to synthesize frames of complex non-linear motions or phenomena, instead of purely relying on the information from the given frames.

Recent advances in image-conditioned diffusion-based text-to-video models~\cite{xing2023dynamicrafter,chen2023seine} suggesting that these large-scale data trained models are capable to synthesize diverse and plausible videos from images.
This motivates us to investigate the feasibility of leveraging the rich motion priors learned in these video models for our generative cartoon interpolation purposes
Unfortunately, directly applying existing models to cartoon interpolation is unsatisfactory due to three factors. 
Firstly, there exists a domain gap as the models are mostly trained on live-action video content. 
Non-cartoon content may be accidentally synthesized. 
The model may also misunderstand the animation-domain content and fails to generate the appropriate motion.
Secondly, to reduce the computational cost, current video diffusion models are based on highly compressed latent spaces~\cite{xing2023dynamicrafter,chen2023seine,rombach2022high}, resulting in significant loss of details and quality degradation. Such degradations are more apparent due to the high-contrast regions, the fine structural outline, and the lack of motion blur in cartoon animations.
Lastly, the generative models can be somewhat random and lack of control. 
An effective control over the generated motion is necessary for cartoon interpolation.

In this paper, we propose an effective and controllable generative framework, \emph{ToonCrafter}, to adapt the pre-trained video diffusion model, and address the three challenges above. ToonCrafter consists of three functional techniques: toon rectification learning, detail injection and propagation in decoding, and sketch-based controllable generation. 
Specifically, the toon rectification learning strategy involves the meticulous fine-tuning of the spatial-related context understanding and content generation layers of the underlying image-conditioned video generation model on collected cartoon data.
This approach offers an accurate context understanding and preserves the rich motion prior while adapting them to the cartoon animation domain. 
To tackle the detail loss and quality degradation issue, we introduce a dual-reference-based 3D decoder, featuring a hybrid-attention-residual-learning mechanism, to convert lossy frame latents back to pixel space. 
It injects the detail information from input images into the generated frame latents using a cross-attention mechanism in shallow decoding layers and residual learning in deeper layers, considering computational cost burden. 
Furthermore, the decoder is equipped with pseudo-3D convolutions to facilitate propagation and improve the temporal coherence.
Lastly, we propose a frame-independent sketch encoder that enables users to flexibly and interactively create or modify interpolation results with temporally sparse or dense motion structure guidance.
As demonstrated in Figure~\ref{fig:teaser}, our ToonCrafter can generate high-quality intermediate frames even in extremely challenging cases with large non-linear motions and dis-occlusions. It also allows users to effectively control the generated motion via sparse sketch input.

We conducted extensive experiments to evaluate the proposed ToonCrafter, which demonstrated considerable superiority over existing competitors. Our contributions are summarized below:
\begin{itemize}
    \item We point out the notion of generative cartoon interpolation and introduce an innovative solution by leveraging live-action video prior. It significantly outperforms existing competitors.
    \item We present a toon rectification learning strategy that effectively adapts live-action motion prior to animation domain.
    \item We propose a dual-reference-based 3D decoder to compensate the lost details resulting from compressed latent space.
    \item Our system enables users to interactively create or modify interpolation results in a flexible and controllable fashion.
\end{itemize}

\section{Related Work}
\label{sec:relatedwork}

\subsection{Video Frame Interpolation}
Video frame interpolation aims at synthesizing multiple frames in between two adjacent frames of the original video, which has been widely studied in recent years. Existing works using deep learning fall into three categories, including phase-based~\cite{meyer2015phase,meyer2018phasenet}, kernel-based~\cite{niklaus2017video,niklaus2017video2,danier2024ldmvfi,zhang2023extracting,jain2024video,zhou2023exploring,liu2023jnmr}, and optical/feature flow-based methods~\cite{huang2022rife,xu2019quadratic,jiang2018super,niklaus2020softmax,xing2021flow,bao2019depth,reda2022film,wu2024perception}. The most recent state-of-the-art has seen more optical flow-based methods, benefited from the latest advancements in flow estimation~\cite{teed2020raft,FlowDiffuser2024,peng2023facial}. The typical approach first identifies the correspondence between two frames using flow, and then performs warping~\cite{glasbey1998review} and fusion. Readers are recommended to refer to~\cite{dong2023video} for a comprehensive survey. Although these methods achieve great success in interpolating live-action videos, they usually fail to handle the large non-linear motion and textureless regions of cartoons.

Existing works tackle part of the above challenges. Zhu et al.~\cite{zhu-2016-toontrack} formulate the cartoon region correspondence as a network flow optimization problem.
AnimeInterp~\cite{siyao2021anime} introduces a segment-guided matching module based on color piece matching, boosting correspondence identification. Furthermore, EISAI~\cite{chen2022eisai} is presented to improve perceptual quality by removing aberrations from solid-color regions using domain-specific perceptual losses. Most recently, Li et al.~\cite{li2021deep} introduced intermediate sketch guidance to address the large motion issues, which, however, is typically not always available due to the necessity of hand drawing. 
Although these methods advance cartoon video interpolation significantly, they rely on the explicit identification of correspondences, as well as the linear or straightfoward motion assumption. They fail to model complex non-linear motion or disocclusion phenomena in cartoon. In contrast, we address these with a new paradigm of generative cartoon interpolation, leveraging the rich live-action video generation prior.

\subsection{Image-conditioned Video Diffusion Models}
%single
Recently, there have been substantial efforts in training large-scale text-to-video (T2V) models~\cite{singer2022make,xing2024make,ho2022imagenvideo,wang2024recipe,blattmann2023align,he2022latent,chen2023videocrafter1,ma2024follow} on large-scale datasets using diffusion models (DMs)~\cite{ho2020denoising}. Moreover, introducing additional image conditions to these T2V models is well-studied for image-to-video (I2V) synthesis~\cite{gu2023seer,wang2024videocomposer,zhang2023i2vgen,xing2023dynamicrafter,blattmann2023stable}. 
Specifically, SEINE~\cite{chen2023seine} is firstly proposed to produce creative transition video segments to connect two different scenes by concatenating two input frames with noisy video latents as input to the diffusion U-Net. 
Similarly, single-image-to-video diffusion models, such as DynamiCrafter~\cite{xing2023dynamicrafter}, SparseCtrl~\cite{guo2023sparsectrl}, and PixelDance~\cite{zeng2023makepixeldance} also exhibit their extensibility for downstream applications of video interpolation/transition by either concatenating two input frames with noisy frame latents~\cite{xing2023dynamicrafter,zeng2023makepixeldance} or using an auxiliary frame encoder~\cite{guo2023sparsectrl}, akin to ControlNet~\cite{zhang2023adding}, respectively. Unfortunately, they are not stable and usable when applying to toon interpolation due to the unique challenges in cartoon.
In this paper, we aim to leverage the rich motion-generative prior in I2V diffusion models learned from live-action videos and adapt it for generative cartoon interpolation.

\section{Method}
\label{sec:method}

\begin{figure}[t]
    \centering
    \includegraphics[width=1\linewidth]{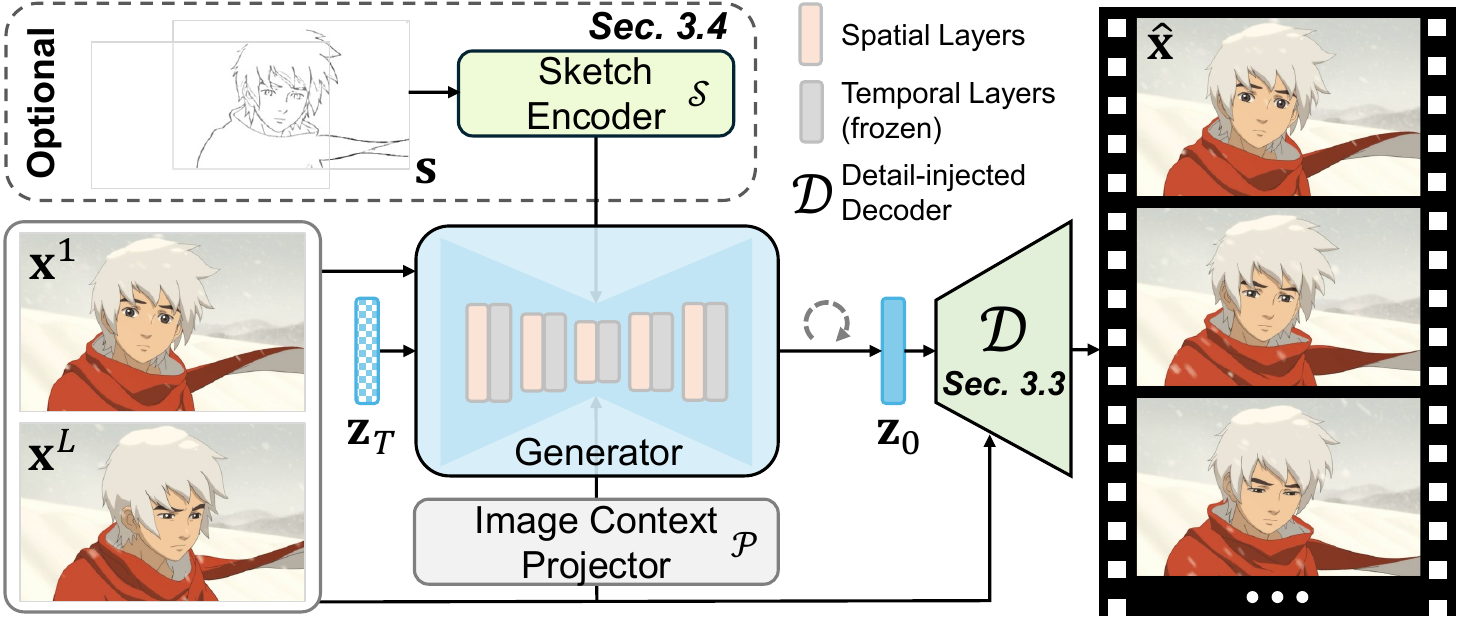}
    \caption{%Flowchart 
    Overview of the proposed \emph{ToonCrafter}. Given two cartoon images $\mathbf{x}^1$ and $\mathbf{x}^L$, ToonCrafter leverages the image-to-video generative diffusion model as a generator to generate intermediate frame latents $\mathbf{z}_0$. These latents are subsequently decoded into pixel space through the proposed detail-injected decoder with $\mathbf{x}^1$ and $\mathbf{x}^L$ as detail guidance. Optionally, the interpolation can be controlled with sparse sketch guidance.
    }
    \label{fig:overview}
    \vspace{-3mm}
\end{figure}

Our generative cartoon interpolation framework is built upon the open-sourced DynamiCrafter interpolation model~\cite{xing2023dynamicrafter}, a state-of-the-art image-to-video generative diffusion model that demonstrates robust motion understanding for live-action interpolation but falls short when applied to cartoon animations. As shown in Figure~\ref{fig:overview}, our framework shares a similar structure to the base model but incorporates three key improvements for generative cartoon interpolation: (1) a meticulously designed toon rectification learning strategy for effective  domain adaptation, (2) a novel dual-reference 3D decoder $\mathcal{D}$ to tackle the visual degradation due to the lossy latent space, and (3) a frame-independent sketch encoder $\mathcal{S}$ that enables the user control over the interpolation.

\subsection{Preliminary}
Diffusion models~\cite{ho2020denoising,sohldickstein2015deep} are score-based generative models that transform data $\mathbf{x}_0 \sim p_{\text{data}}(\mathbf{x})$ into Gaussian noise $\mathbf{x}_T \sim \mathcal{N}(\mathbf{0}, \mathbf{I})$ via a forward diffusion process and learn to reverse it through denoising. The forward process $q(\mathbf{x}_t|\mathbf{x}_0,t)$ has $T$ timesteps, progressively adding noise to the data sample $\mathbf{x}_0$ to produce $\mathbf{x}_t$ using a parameterization trick. The denoising process $p_{\theta}(\mathbf{x}_{t-1}|\mathbf{x}_t,t)$ retrieves less noisy data $\boldsymbol{x}_{t-1}$ from the noisy input $\boldsymbol{x}_t$ via a denoising network $\mathbf{\epsilon}_{\theta}\left(\mathbf{x}_t, t\right)$, with the objective: 
\begin{equation}
\min_{\theta} \mathbb{E}_{t, \mathbf{x} \sim p_{\text{data}}, \mathbf{\epsilon} \sim \mathcal{N}(\mathbf{0}, \mathbf{I})}\Vert\mathbf{\epsilon}-\mathbf{\epsilon}_{\theta}\left(\mathbf{x}_t, t\right)\Vert_2^2, 
\end{equation} 
where $\theta$ denotes learnable network parameters and $\mathbf{\epsilon}$ represents the sampled ground truth noise. After training, denoised data $\mathbf{x}_0$ can be obtained from random noise $\mathbf{x}_T$ via iterative denoising.

In the video generation realm, Latent Diffusion Models (LDMs)~\cite{rombach2022high} are widely employed to reduce computational complexity. In this work, we base our study on an image-to-video generation LDM, \emph{DynamiCrafter}~\cite{xing2023dynamicrafter}. Given a video $\mathbf{x} \in \mathbb{R}^{L\times 3\times H\times W}$, we first encode it into a latent representation $\mathbf{z}=\mathcal{E}(\mathbf{x}), \mathbf{z} \in \mathbb{R}^{L\times C\times h\times w}$ on a frame-by-frame basis. Next, both the forward diffusion process $\mathbf{z}_t=p(\mathbf{z}_0, t)$ and backward denoising process $\mathbf{z}_t=p_{\theta}(\mathbf{z}_{t-1}, \mathbf{c}, t)$ are executed in this latent space, where $\mathbf{c}$ represents the denoising conditions such as text $\mathbf{c}_\text{txt}$ and image prompts $\mathbf{c}_\text{img}$.
Following the description in DynamiCrafter, the interpolation application is realized by providing both starting and ending frames $\mathbf{x}^1$ and $\mathbf{x}^L$ while leaving middle frames as empty for $\mathbf{c}_\text{img}$. Then, the objective is:
\begin{equation}
% \scalebox{.8}{$\displaystyle
\min_{\theta} \mathbb{E}_{\mathcal{E}(\mathbf{x}), t, \mathbf{\epsilon} \sim \mathcal{N}(\mathbf{0}, \mathbf{I})} \left[\Vert\mathbf{\epsilon}-\mathbf{\epsilon}_{\theta}\left(\mathbf{z}_t;\mathbf{c}_\text{img},\mathbf{c}_\text{txt},t,fps\right)\Vert_2^2 \right],
%$}
% \vspace{-1em}
\end{equation}
where $fps$ is the FPS control introduced in~\cite{xing2023dynamicrafter}. The generated videos are then obtained through the decoder $\hat{\mathbf{x}}=\mathcal{D}(\mathbf{z}_0)$.

%%%%%%%%%%%%%%%%%%%%%%%%%%%%%%%%%%%%%%%%%%%%%%%%%
%%%%%%%%%%%%%%%%%%%%%%%%%%%%%%%%%%%%%%%%%%%%%%%%%

\subsection{Toon Rectification Learning}
\label{sec:toon-learning}
Cartoon and live-action videos exhibit a domain gap due to factors such as distinct visual styles, exaggerated expressions, and simplified textures. 
This domain gap poses challenges to applying existing video generative prior models (mainly trained on live-action videos) to cartoon animations. 
Some potential issues include the unintentional synthesis of non-cartoon content (Figure~\ref{fig:ablation_toon_rect}-I), as well as the model's inability to accurately comprehend animation content, leading to the generation of inappropriate motion.
To address this, we first collect a cartoon video dataset and then adapt the motion prior model to the cartoon domain by meticulously designed fine-tuning.

\textit{Cartoon Video Dataset Construction.} 
We collect a series of raw cartoon videos and then manually select high-quality ones based on %the bit rate, 
the resolution and subjective quality. The total duration of the selected videos is more than 500 hours.
We employ PySceneDetect~\cite{pyscenedetect} to detect and split shots.
The static shots are filtered out by removing any videos with low average optical flow~\cite{teed2020raft} magnitude.
Moreover, we apply optical character recognition (CRAFT)~\cite{baek2019character} to
weed out clips containing large amounts of text.
In addition, we adopt LAION~\cite{schuhmann2021laion} regressive model to calculate the aesthetic score for removing the low-aesthetic samples to ensure quality.
Next, we annotate each clip with the synthetic captioning method BLIP-2~\cite{li2023blip}.
Lastly, we annotate the first, middle, and last frames of each video clip with CLIP~\cite{radford2021learning} embeddings from which we measure the text-video alignment, to filter out mismatched samples.

In the end, we obtained 271K high-quality cartoon video clips, which were randomly split into two sets. The training set contains 270K clips, while the evaluation set contains 1K clips.  

\textit{Rectification Learning.} 
With the collected cartoon text-video data, we can then adapt the DynamiCrafter interpolation model (DCinterp) trained on live-action videos for cartoon interpolation. However, directly fine-tuning the denoising network of DCinterp on our data would lead to catastrophic forgetting due to \textbf{unbalanced scale} between our cartoon video data (\textit{270K} video clips) and the original training data of DCinterp (WebVid-10M \cite{Bain21}, \textit{10M}), which deteriorates motion prior, as evidenced in Sec.~\ref{subsec:ablation_study}. 
One straightforward solution is to expand our training set by orchestrated allocation of the original training data of DCinterp and our fine-tuning data, but this approach would require significantly more training compute. To address this issue, we design an efficient rectification learning strategy that allows for fine-tuning the base model using only a small-scale cartoon dataset without compromising the robust motion prior obtained from the large-scale live-action videos. 

The DCinterp model consists of three key components: an image-context projector, the spatial layers (sharing the same architecture as StableDiffusion v2.1), and the temporal layers. 
Based on our experiments (Sec.~\ref{sec:exp:albate:rectification}), we have the following observations: the image-context projector helps the DCinterp model to digest the context of the input frames; the spatial layers are responsible for learning the appearance distribution of video frames; the temporal layers capture the motion dynamics between the video frames.

In other words, the temporal layers should be frozen during the fine-tuning to preserve the real-world motion prior as illustrated in Figure~\ref{fig:overview}.  On the other hand,
the image-context projector can be fine-tuned to achieve better semantic alignment and allow the model to digest the cartoon scene context better. Simultaneously, spatial layers should also be tuned for appearance rectification, thereby preventing the generation of live-action video content in the intermediate frames.  
Even though the cartoon animation might exhibit motions (not as rigid as possible for cases) that are slightly different from the real-world motions, the high-level motion concepts are still the same (otherwise, human viewers cannot recognize what the motion is), making the appearance the dominant factor in domain adaptation for cartoon animation. In summary, our toon rectification learning strategy focuses on the appearance by freezing the temporal layers (to preserve the real-world motion prior) and finetuning the image-context projector and spatial layers with only our collected cartoon data to achieve effective domain adaptation.

%%%%%%%%%%%%%%%%%%%%%%%%%%%%%%%%%%%%%%%%%%%%%%%%%
\begin{figure}[t]
    \centering
    \includegraphics[width=\linewidth]{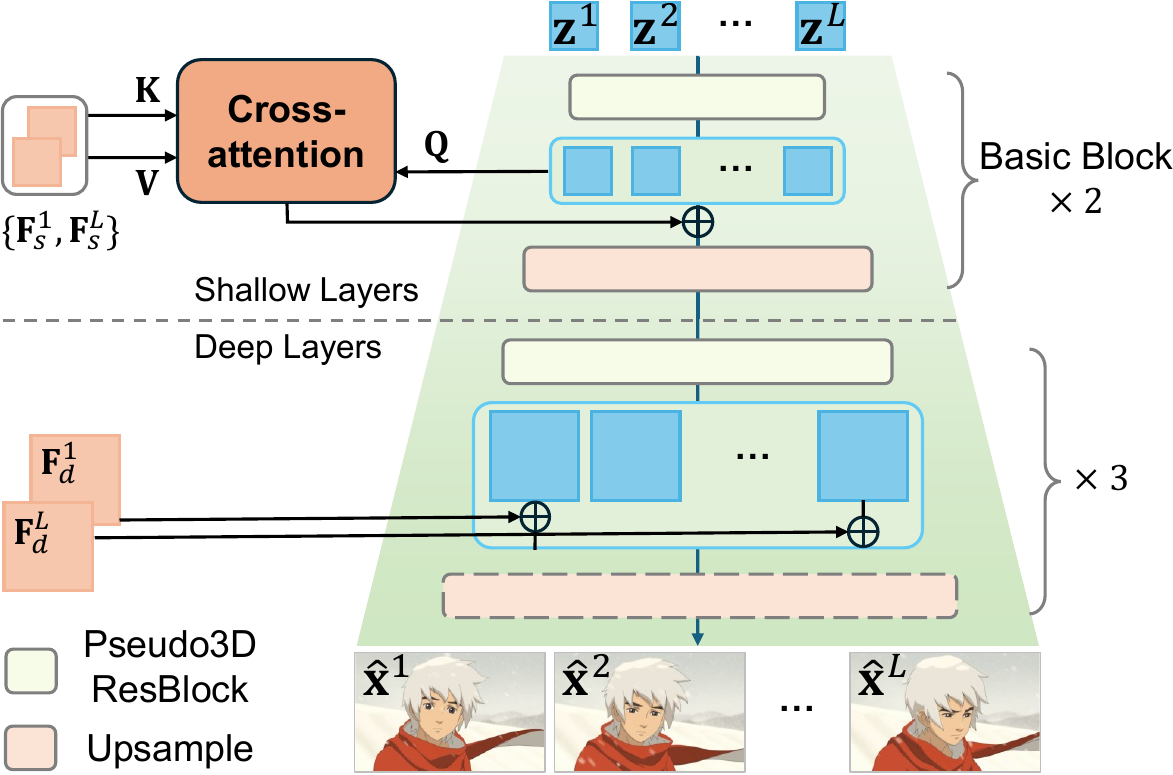}
    % \vspace{-0.5em}
    \caption{
    Illustration of the detail-injected 3D decoder. Given frame latents $\mathbf{z}$ as input, we inject the intermediate features of input images $\mathbf{x}^1$ and $\mathbf{x}^L$ from encoder $\mathcal{E}$ through cross-attention in shallow layers, while via residual learning, \ie, addition to features of 1-st and $L$-th frame in deep layers.
    }
    \label{fig:decoder}
    % \vspace{-1em}
\end{figure}
%%%%%%%%%%%%%%%%%%%%%%%%%%%%%%%%%%%%%%%%%%%%%%%%%
\subsection{Detail Injection and Propagation in Decoding} \label{sec:decoder}
Most of the current video diffusion models~\cite{blattmann2023align,wang2024videocomposer}, 
including DynamiCrafter~\cite{xing2023dynamicrafter} we built upon, 
learn to generate the videos in highly compressed latent spaces~\cite{rombach2022high}, which are typically obtained through vector quantized auto-encoding (VQ-VAE)~\cite{vq-gan}.
The latent video diffusion paradigm effectively reduces computational demand. However, it results in a significant loss of details and intolerable quality degradation including flickering and distortion artifacts. Unfortunately, such degradations are more apparent in cartoon animation due to its typical appearance of high-contrast regions, fine structural outline, and the lack of motion blur (motion blur in live-action video effectively ``hides'' the degradation). 
Consequently, directly applying the latent video diffusion model for cartoon interpolation could lead to unacceptable results, as the structure and texture are highly likely to contain artifacts and be inconsistent with the original input frames (4-th row in Figure~\ref{fig:decoder_vis}).
While several solutions try to improve the decoding quality, they either require perfectly-aligned references~\cite{liu2023video,liu2023improved} or only work for single frame generation 
% single image generation 
\cite{danier2024ldmvfi}, or sacrifice high-frequency components in the results \cite{blattmann2023stable}.

% overall idea
To address this issue, we propose to exploit the existing information from the input frames and introduce a dual-reference-based 3D decoder to propagate the pixel-level details from the two input frames to the decoding process of the generated lossy-space frame latents.
Rather than relying solely on the decoder $\mathcal{D}$ to recover the compressed details, we firstly extract the inner features $\{\mathbf{F}^K_i\}$ at each residual block of $\mathcal{E}$ (where $i$ represents the $i$-th residual block from the end in the encoder and $K$ indicates the $K$-th frame), and then inject them into the decoding process. This provides the necessary hints for achieving pixel-perfect compensation.
Specifically, we propose a hybrid-attention-residual-learning mechanism (HAR) to inject and propagate details. As shown in Figure~\ref{fig:decoder}, we introduce cross-frame-attention in $\mathcal{D}$ to inject the intricate details from $\{\mathbf{F}_i^1\}_{i\in s}$ and $\{\mathbf{F}_i^L\}_{i\in s}$ to decoder's intermediate features $\mathbf{G}_{\text{in}}$:
%for each frames:
\begin{equation}
    \mathbf{G}^{j}_{\text{out}} = \text{Softmax}(\frac{\mathbf{Q} \mathbf{K}^{\top}}{\sqrt{d}})\mathbf{V} + \mathbf{G}^{j}_{\text{in}}, j \in 1...L 
\end{equation}
where $\mathbf{Q}=\mathbf{G}^{j}_{\text{in}}\mathbf{W}_\mathbf{Q}$, $\mathbf{K}=[\mathbf{F}_i^1;\mathbf{F}_i^L]\mathbf{W}_\mathbf{K}$, $\mathbf{V}=[\mathbf{F}_i^1;\mathbf{F}_i^L]\mathbf{W}_\mathbf{V}$, and $[;]$ denotes concatenation.
% where $\mathbf{Q}=\mathbf{G}_{\text{in}}\mathbf{W}_\mathbf{Q}$, $\mathbf{K}=[\mathbf{F}_i^1;\mathbf{F}_i^L]\mathbf{W}_\mathbf{K}$, $\mathbf{V}=[\mathbf{F}_i^1;\mathbf{F}_i^L]\mathbf{W}_\mathbf{V}$, and $[\cdot]$ denotes concatenation.
Considering the computational cost of attention, we implement it only in the first two layers (\ie, shallow layers $s=\{1,2\}$) of $\mathcal{D}$. 
Since the input frame $\mathbf{x}^1$ and resultant $\hat{\mathbf{x}}^1$ are aligned in pixel-level, we add the \texttt{ZeroConv}~\cite{zhang2023adding} processed $\{\mathbf{F}_i^1\}_{i\in d}$ (\texttt{ZeroConv} processed $\{\mathbf{F}_i^L\}_{i\in d}$) to the corresponding feature maps of the first frame (the $L$-th frame),
\begin{equation}
    \mathbf{G}_{\text{out}}^1 = \texttt{ZeroConv}_{1\times1}(\mathbf{F}_i^1) + \mathbf{G}_{\text{in}}^1.
\end{equation}
To avoid redundant computation, we implement this residual learning only at the deep layers ($d=\{3,4,5\}$) of $\mathcal{D}$.  
In addition, we incorporate pseudo-3D convolutions (P3D) to further facilitate the propagation and improve the temporal coherence.

\noindent\textbf{Training.} 
We freeze the image encoder $\mathcal{E}$ and optimize the proposed decoder $\mathcal{D}$, which is initialized from the vanilla image decoder.
We use a compound loss $\mathcal{L}$ to encourage reconstruction:
\begin{equation}
\mathcal{L}=\mathcal{L}_1+\lambda_{p}\mathcal{L}_p+\lambda_{d}\mathcal{L}_d,
\end{equation}
where $\mathcal{L}_1$ is the MAE loss, $\mathcal{L}_p$ is a perceptual loss (LPIPS \cite{zhang2018unreasonable}), $\mathcal{L}_d$ is a discriminator loss~\cite{hingeloss}, and $\lambda_p=0.1$, $\lambda_d$ is an adaptive weight following~\cite{vq-gan}.

%%%%%%%%%%%%%%%%%%%%%%%%%%%%%%%%%%%%%%%%%%%%%%%%%
\begin{figure}[t]
    \centering
    \includegraphics[width=0.95\linewidth]{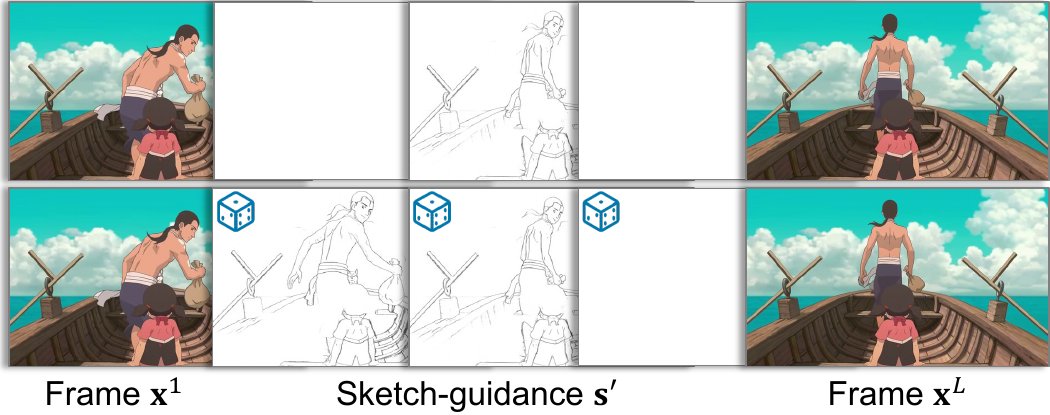}
    \vspace{-2.5mm}
    \caption{
    Examples of different patterns of sketch-guidance: (top) bisection ($n$=1) and (bottom) random position.
    }
    \label{fig:sketch_encoder}
    \vspace{-2.5mm}
\end{figure}
% index=4,11,18 for the three sketches
%%%%%%%%%%%%%%%%%%%%%%%%%%%%%%%%%%%%%%%%%%%%%%%%%
\subsection{Sketch-based Controllable Generation}
\label{sec:sketch-encoder}

The generative nature of our framework enables nonlinear motion interpolation but unavoidably introduces variations in the generation results. While such variation is needed for some users, others may prefer better control over the frame interpolation.
To make our framework more controllable for real-world production settings, 
we follow the industrial practices \cite{li2021deep} and introduce %more flexible 
sketch-based generation guidance.
We propose a frame-independent sketch encoder $\mathcal{S}$ that enables users to control the generated motion using sparse sketch guidance. 
Built upon the adapter mechanism \cite{zhang2023adding}, our sketch encoder effectively converts our video diffusion model into a sketch-conditioned generative model.

To reduce users' drawing efforts and accommodate real-world usage, we design our sketch encoder module $\mathcal{S}$ that supports sparse inputs (Figure~\ref{fig:sketch_encoder}) in which the users are not required to provide all sketch images for the target frames.
To achieve this, we design $\mathcal{S}$ as a frame-wise adapter that learns to adjust the intermediate features of \textit{each frame independently} based on the provided sketch:
$
\mathbf{F}_\text{inject}^{i}=\mathcal{S}(\mathbf{s}^i,\mathbf{z}^i,t),
$
where $\mathbf{F}_\text{inject}^{i}$ is processed using the same strategy of ControlNet \cite{zhang2023adding}. 
For the frames without sketch guidance, $\mathcal{S}$ takes an empty image as input: 
$
\mathbf{F}_\text{inject}^{i}=\mathcal{S}(\mathbf{s}^\emptyset,\mathbf{z}^i,t).
$
Using empty image inputs improves the learning dynamics of sketch encoder. If we simply omit the guidance for frames without sketch input, the back-propagated gradient is  only from the frames with user-provided sketches, where the adapter could only learn to control the spatial contents of a certain frame but compromise the temporal consistency across the frame sequence. 

\noindent\textbf{Training.} We freeze the denoising network $\mathbf{\epsilon}_{\theta}$ and optimize the sketch encoder $\mathcal{S}$. $\mathcal{S}$ uses a ControlNet-like architecture, initialized from the pre-trained StableDiffusion v2.1. The training objective is:  
\begin{equation}
% \scalebox{.8}{$\displaystyle
\min_{\theta} \mathbb{E}_{\mathcal{E}(\mathbf{x}), \mathbf{s}, t, \mathbf{\epsilon} \sim \mathcal{N}(\mathbf{0}, \mathbf{I})} \left[\Vert\mathbf{\epsilon}-\mathbf{\epsilon}_{\theta}^{\mathcal{S}}\left(\mathbf{z}_t;\mathbf{c}_\text{img},\mathbf{c}_\text{txt}, \mathbf{s}',t,fps\right)\Vert_2^2 \right],
%$}
% \vspace{-1em}
\end{equation}
where $\mathbf{\epsilon}_{\theta}^{\mathcal{S}}$ denotes the combination of $\mathbf{\epsilon}_{\theta}$ and $\mathcal{S}$, $\mathbf{s}$ denotes sketches obtained from Anime2Sketch\cite{Anime2Sketch} using original video frames, and $\mathbf{s}'$ denotes selected sketches from $\mathbf{s}$ (illustrated in Figure~\ref{fig:sketch_encoder}). 
To support typical patterns of user-provided sketch inputs, we design a bisection selection pattern (chosen 80\% of the time) to select input sketches: for an interpolation segment $(i,j)$, the sketch of $\floor{\frac{i+j}{2}}$-th frame is selected; the selection is applied recursively (with the  recursion depth $n$ uniformly sampled from $[1,4]$) from segment $(1,L)$ to subdivided segments. 
This bisection selection pattern mimics real-world user behavior, where users provides sketches at equally-spaced interval. For the remaining 20\%, we randomly select input sketches from $\mathbf{s}$ to maximize the generalization ability. 
\section{Experiments}
\label{sec:experiments}
\subsection{Implementation Details}
Our implementation is primarily based on the image-to-video model DynamiCrafter~\cite{xing2023dynamicrafter} (interpolation variant @512$\times$320 resolution). In the toon rectification learning, we trained spatial layers and $\mathcal{P}$ with 50K steps on the learning rate (lr) of $1\times10^{-5}$ and a mini-batch size (bs) of 32. We trained our dual-reference-based 3D decoder with 60K steps on the lr=$4.5\times10^{-6}$ and bs=16. The sketch encoder was trained with 50K on the lr=$5\times10^{-5}$ and bs=32. Our ToonCrafter was trained on the collected cartoon video dataset by sampling 16 frames with dynamic FPS at the resolution of 512$\times$320 in a batch. At inference, we adopt DDIM sampler~\cite{song2021denoising} with classifier-free guidance~\cite{ho2022classifier}.

%%%%%%%%%%%%%%%%%%%%%%%%%%%%%%%%%%%%%%%%%%%%%%%%
\begin{table}[t]
  \caption{Quantitative comparison with state-of-the-art video interpolation methods on the cartoon test set (1K).}
  \vspace{-0.5em}
  \label{tab:quan_compare}
\resizebox{\linewidth}{!}{
  \centering
  \begin{tabular}{lcccca}
  \toprule
    Metric & AnimeInterp &EISAI &FILM &SEINE &\mc{1}{\multirow{1}{*}{Ours}}\\
    \midrule
FVD $\downarrow$ &196.66 &146.65 &189.88 &98.96 &\best{43.92}\\
 KVD $\downarrow$ &8.44 &5.55 &8.01 &2.93 &\best{1.52}\\
 \midrule
 LPIPS $\downarrow$& 0.1890&0.1729 &\best{0.1702} & 0.2519&0.1733\\
 CLIP$_\text{img}$ $\uparrow$ &0.8866 & 0.9083& 0.9006& 0.8531&\best{0.9221}\\
 CLIP$_\text{txt}$ $\uparrow$ &0.3069 & 0.3097&0.3083 &0.2962 &\best{0.3129}\\
 \midrule
 CPBD $\uparrow$& 0.5974& 0.6413&0.6317 & 0.6630&\best{0.6723}\\
  \bottomrule
  \end{tabular}
}
\vspace{-0.5em}
\end{table}
%%%%%%%%%%%%%%%%%%%%%%%%%%%%%%%%%%%%%%%%%%%%%%%%

\subsection{Quantitative Comparisons}

\paragraph{Metrics and datasets.} 
To evaluate the quality and temporal motion dynamics of generated videos in both the spatial and temporal domains, we report Fr\'echet Video Distance (FVD)~\cite{unterthiner2019fvd} as well as Kernel Video Distance (KVD). In addition, we conduct image-level evaluation. Following~\cite{chen2022eisai}, we employ LPIPS~\cite{zhang2018unreasonable} to measure the perceptual similarity with the ground-truth videos. To evaluate the correctness of semantics in the generated video frames, we calculate cosine similarity between the CLIP~\cite{radford2021learning} features of generated frames and the GT frames \& text prompts, separately, and denote them as CLIP$_\text{img}$ and CLIP$_\text{txt}$, respectively. We employ cumulative probability blur detection (CPBD)~\cite{narvekar2011no} to evaluate sharpness.
We evaluate these metrics on the 1K evaluation set of our cartoon video dataset (@512$\times$320 with 16 frames).

We compare our method against various representative state-of-the-arts: cartoon video interpolation methods (AnimeInterp~\cite{siyao2021anime} and EISAI~\cite{chen2022eisai}), general video interpolation method for handling large motion (FILM~\cite{reda2022film}), and generative video transition (SEINE~\cite{chen2023seine}), which are fine-tuned on our dataset for fair comparison. The quantitative results are presented in Table~\ref{tab:quan_compare}. According to the results, our proposed method significantly outperforms previous approaches in almost all metrics except for LPIPS. However, we argue that LPIPS is a full-reference metric and requires pixel-level alignment, which is less suitable in the generative context, as the ground-truth intermediate frames might not be the only correct answer, especially for ambiguous motions with multiple valid outcomes.

\subsection{Qualitative Comparisons}
\begin{figure*}[t]
    \centering
    \includegraphics[width=0.96\linewidth]{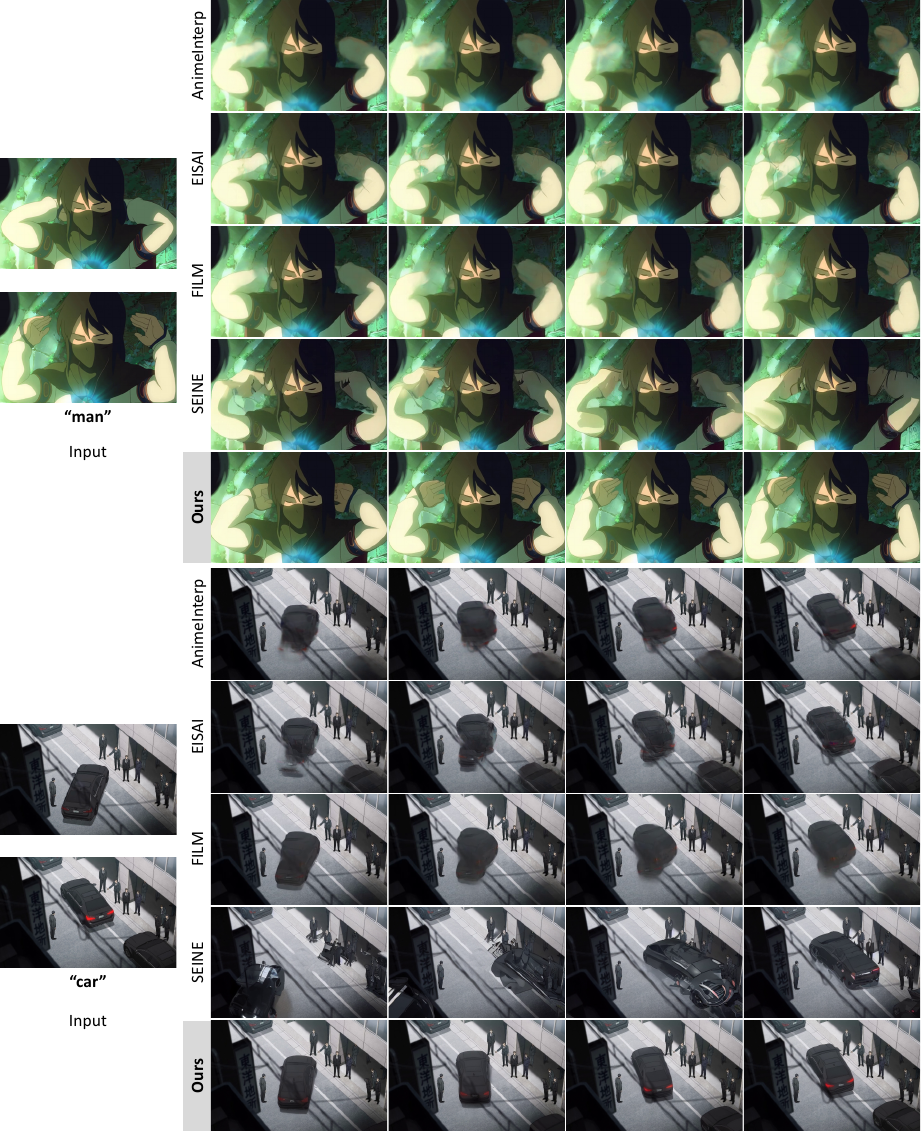}
    \caption{
    Visual comparison of cartoon interpolation results from AnimeInterp, EISAI, FILM, SEINE, and our ToonCrafter.
    }
    \label{fig:qualitative_compare}
    % \vspace{-3mm}
\end{figure*}

The visual comparisons of representative interpolation results are shown in Figure~\ref{fig:qualitative_compare}. Among all compared methods, our approach generates intermediate frames with non-linear plausible motions. In contrast, traditional correspondence-based methods (\ie, AnimeInterp, EISAI, and FILM) struggle to produce natural motions due to their failure to handle dis-occlusions (`dissolving' hands and arms in `man' case of Figure~\ref{fig:qualitative_compare}). These methods also struggle with synthesizing complex non-linear motions (morphing in `car' case of Figure~\ref{fig:qualitative_compare}). While SEINE can potentially generate videos exhibiting decent motions, it suffers from severe distortions in both structure and texture. More videos are provided in the \emph{Supplement} for better comparisons.

\subsection{User Study}
We conduct a user study (detailed in the supplementary) to evaluate the perceptual quality of the results. The participants were asked to choose the best result in terms of motion quality, temporal coherence, and frame fidelity. The statistics from the responses of 24 participants are presented in Table~\ref{tab:user_study}. Our method demonstrates significant superiority over other competitors. 
%%%%%%%%%%%%%%%%%%%%%%%%%%%%%%%%%%%%%%%%%%%%%%%%
\begin{table}[t]
  \caption{User study statistics of the preference rate for Motion Quality (M.Q.), Temporal Coherence (T.C.), and Frame Fidelity (F.F.).}
  \vspace{-0.5em}
  \label{tab:user_study}
\resizebox{\linewidth}{!}{
  \centering
  \begin{tabular}{lcccca}
  \toprule
    Property & AnimeInterp &EISAI &FILM &SEINE &\mc{1}{\multirow{1}{*}{Ours}}\\
    \midrule
    M.Q. $\uparrow$& 3.24\%& 6.94\%& 6.02\%& 14.81\%&\best{68.98}\%\\
    T.C. $\uparrow$& 6.94\%&14.81\% &13.43\% & 15.74\%&\best{49.07}\%\\
    F.F. $\uparrow$& 6.48\%&11.57\% &12.50\% &18.06\% &\best{51.39}\%\\
  \bottomrule
  \end{tabular}
}
\vspace{-0.5em}
\end{table}
%%%%%%%%%%%%%%%%%%%%%%%%%%%%%%%%%%%%%%%%%%%%%%%%

\subsection{Ablation Study}
\label{subsec:ablation_study}

\begin{figure}[t]
    \centering
    \includegraphics[width=1\linewidth]{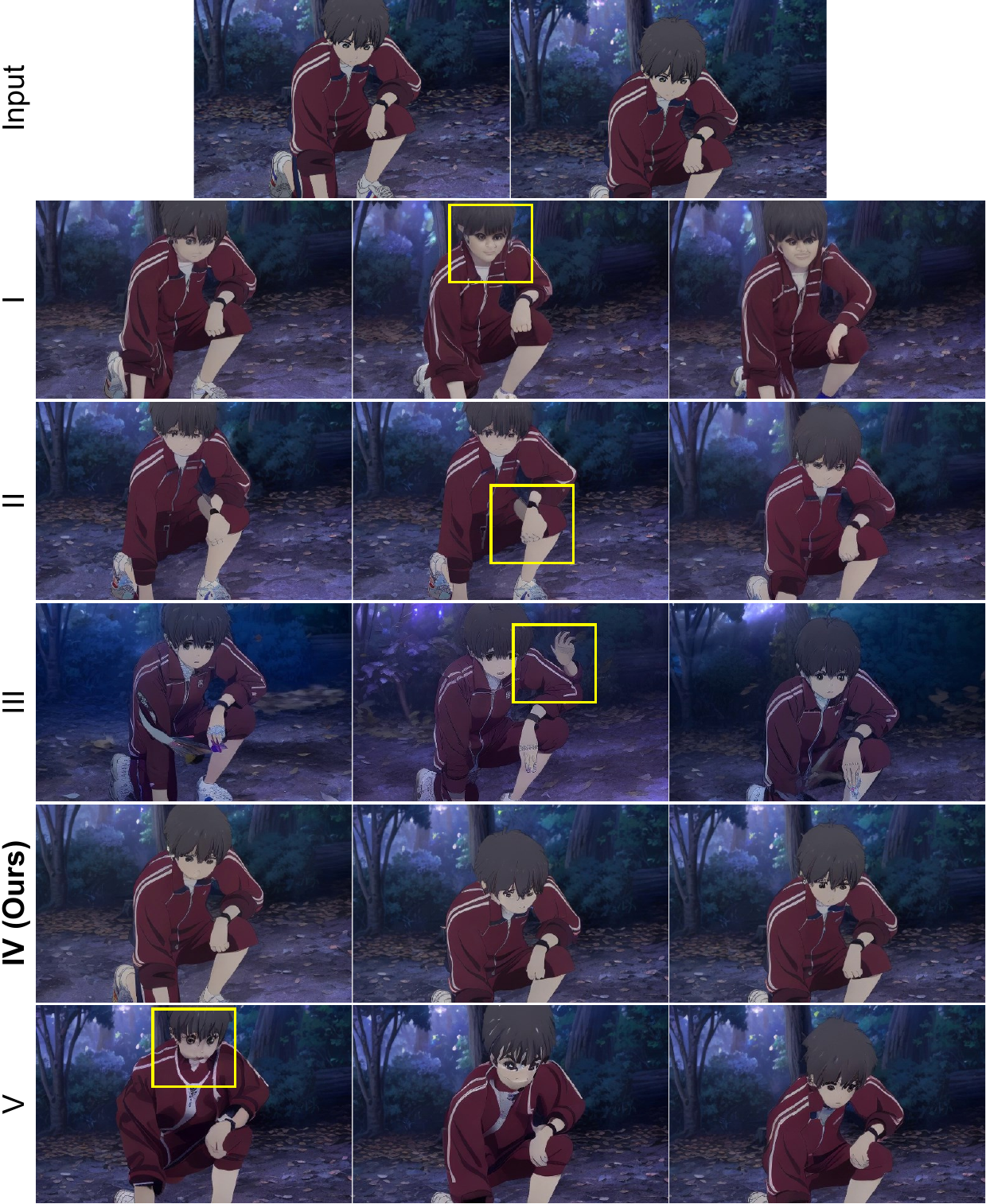}
    \vspace{-1.5em}
    \caption{
    Visual comparison of the interpolation frames generated by variants with different rectification strategies.
    }
    \label{fig:ablation_toon_rect}
    \vspace{-0.5em}
\end{figure}

\textbf{\textit{Toon rectification learning.}}
\label{sec:exp:albate:rectification}
We construct the following baselines to investigate the effectiveness of our domain adaptation strategy:
\textbf{(I)}: directly using the pre-trained backbone model (DCinterp~\cite{xing2023dynamicrafter}), 
\textbf{(II)}: fine-tuning the image-context projector (ICP) and entire denoising U-Net (Spatial+Temporal layers),
\textbf{(III)}: fine-tuning ICP and spatial layers while bypassing temporal layers in forwarding during training,
\textbf{(IV)} (our adapting strategy): fine-tuning ICP and spatial layers while keeping temporal layers frozen ,
and \textbf{(V)}: fine-tuning only ICP. The quantitative comparison is shown in Table~\ref{tab:ablation_toon_rect}. DCInterp without any fine-tuning (I) shows decent quantitative results but suffers from unexpected generation of live-action content (2nd row in Figure~\ref{fig:ablation_toon_rect}). While directly fine-tuning all layers (II) leads to adaption to cartoon domain to some extent, it deteriorates the temporal prior, as evidenced by the inconsistency and sudden change of generated content (3rd row in Figure~\ref{fig:ablation_toon_rect}). Moreover, simply bypassing temporal layers (III) in forwarding to preserve temporal prior leads to disastrous degradation due to mismatched distribution mapping.
Comparing (I), (II), and (IV), we can observe improved performance of both FVD and CLIP$_\text{img}$ by fine-tuning ICP and spatial layers, while keeping temporal layers frozen, which enhances the adaptability to cartoon domain and preserves learned motion prior. The comparison between (I) and (V) shows fine-tuning ICP slightly improves semantic alignment for generating semantically correct content (higher CLIP$_\text{img}$), thanks to its better comprehension of cartoon context.

%%%%%%%%%%%%%%%%%%%%%%%%%%%%%%%%%%%%%%%%%%%%%%%%
\begin{table}[!t]
  \caption{Ablation study of different rectification learning strategies. $^*$All these variants are evaluated without the proposed dual-reference-based 3D decoder to demonstrate the original performance of the denoising network.}
  \vspace{-0.5em}
  \label{tab:ablation_toon_rect}
\resizebox{\linewidth}{!}{
  \setlength{\tabcolsep}{2.5pt}
  \centering
  \begin{tabular}{lcccc|cc}
  \toprule
    Variant$^*$  & ICP & Spa. &Temp. & Bypass Temp. & FVD $\downarrow$ & CLIP$_\text{img}$ $\uparrow$ \\
    \midrule
    I (DCinterp)& & & & & 86.62&0.8637\\
    II& \cmark&\cmark &\cmark & & 70.73&0.8978\\
   III &\cmark &\cmark & &\cmark &291.45& 0.7997\\
    \rowcolor{Gray}
    IV (Ours)& \cmark&\cmark & & &\best{52.73}& \best{0.9096}\\
    V& \cmark& & & &81.45&0.8875\\
  \bottomrule
  \end{tabular}
}
  \vspace{-0.5em}
\end{table}
%%%%%%%%%%%%%%%%%%%%%%%%%%%%%%%%%%%%%%%%%%%%%%%%
%%%%%%%%%%%%%%%%%%%%%%%%%%%%%%%%%%%%%%%%%%%%%%%%%
%%%%%%%%%%%%%%%%%%%%%%%%%%%%%%%%%%%%%%%%%%%%%%%%%

\begin{figure}[t]
    \centering
    \includegraphics[width=1\linewidth]{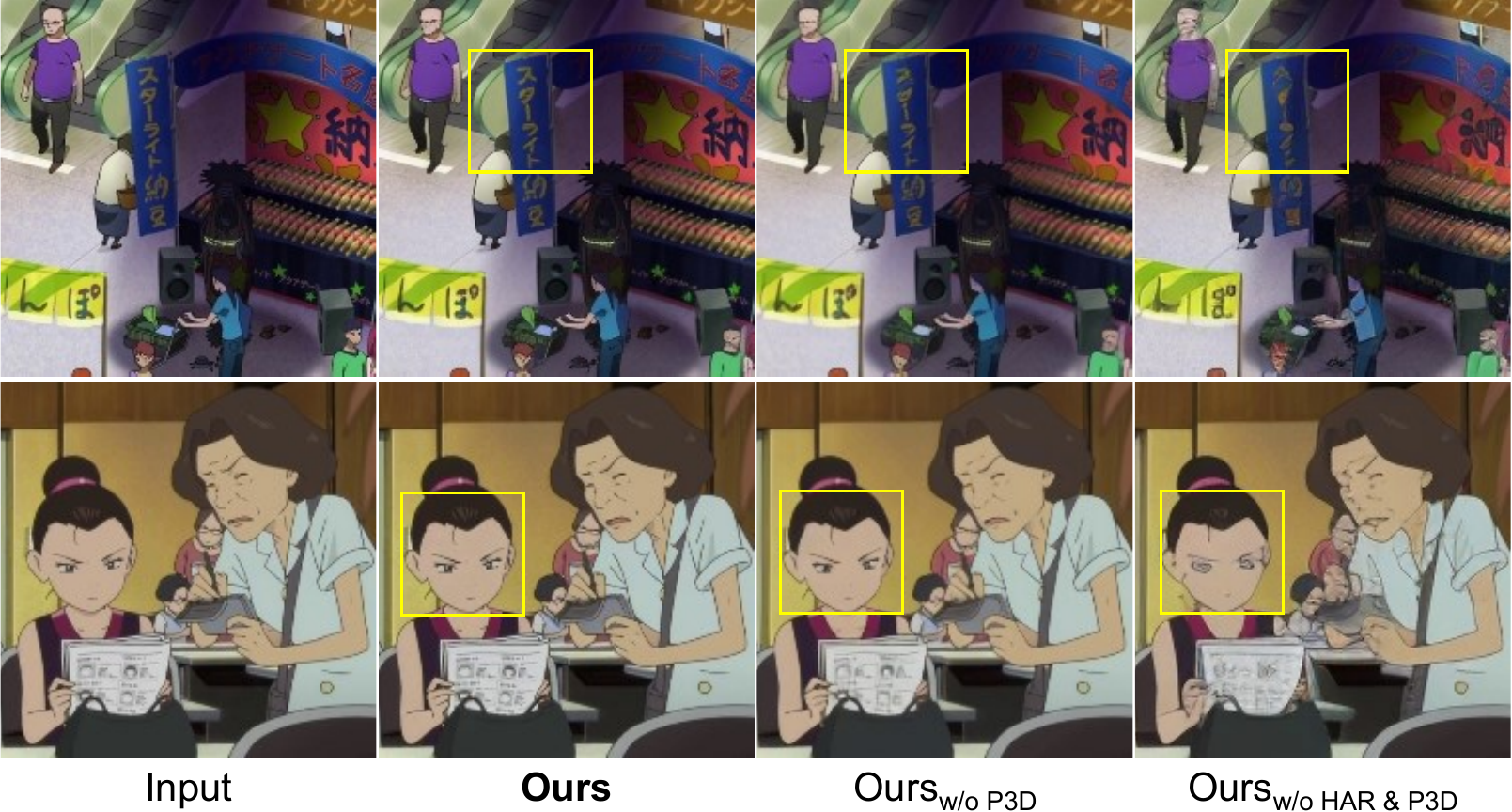}
    \vspace{-0.5em}
    \caption{
    Visual comparison of the reconstructed video frames produced by different decoder variants. Only the middle frames are shown above.
    }
    \label{fig:decoder_vis}
    \vspace{-0.5em}
\end{figure}

\begin{figure}[t]
    \centering
\includegraphics[width=0.85\linewidth]{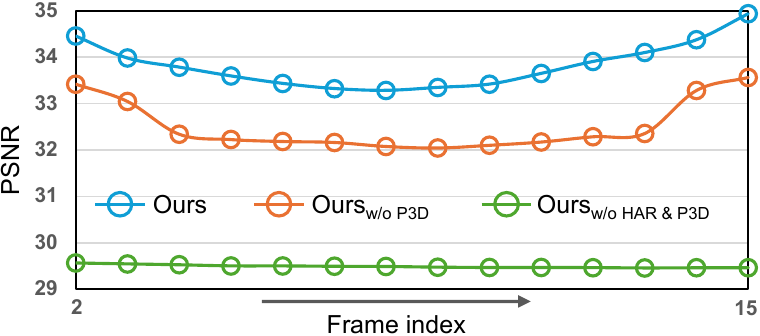}
\vspace{-0.7mm}
    \caption{
    Comparison of reconstruction quality against the frame index.
    }
    \label{fig:decoder_stat}
    \vspace{-0.8em}
\end{figure}

\textbf{\textit{Dual-reference-based 3D VAE decoder.}}
We further evaluate the effectiveness of different modules in the proposed dual-reference-based 3D decoder. We first construct a variant by removing the pseudo-3D convolutions (P3D), denoted by \textbf{Ours$_\text{w/o P3D}$}. Base on that, we then further remove the introduced hybrid-attention-residual (HAR) module to obtain the second variant \textbf{Ours$_\text{w/o HAR \& P3D}$}, which is exactly the image decoder used in most diffusion-based image/video generative models. 
We evaluate the our full method with the mentioned variant via video reconstruction task and report the evaluation results in Table~\ref{tab:decoder}.
The performance of `Ours$_\text{w/o P3D}$' declines due to the attenuation of propagation for injected reference information, leading to a loss of details (3rd column in Figure~\ref{fig:decoder_vis}).
Furthermore, removing both HAR and P3D considerably impairs the performance, as shown in Table~\ref{tab:decoder}, since solely relying on the VAE decoder without reference fails to recover lost details in compressed latents (4th column in Figure~\ref{fig:decoder_vis}). In contrast, our full method effectively compensates for the lost details through the introduced dual-reference-based detail injection and propagation mechanism. 
Additionally, we show the comparison of reconstruction quality along frame sequences (average on the 1K test samples) in Figure~\ref{fig:decoder_stat}, which further highlights the effectiveness of our design. 

%%%%%%%%%%%%%%%%%%%%%%%%%%%%%%%%%%%%%%%%%%%%%%%%
\begin{table}[t]
% \vspace{-2mm}
% \setlength{\tabcolsep}{3.2pt}
\centering
\caption{
  Quantitative comparison of reconstruction results from different decoders on the 1K cartoon video evaluation set (256$\times$256). HAR: Hybrid-attention-residual. P3D: Pseudo-3D Convolution.
}
\vspace{-0.5em}
\resizebox{\linewidth}{!}{
\begin{tabular}{lcc|cccc}
  \toprule
  Variant& Ref. & Temp.   & PSNR $\uparrow$ & SSIM $\uparrow$ & LPIPS $\downarrow$\\
  \midrule
  \rowcolor{Gray}
    Ours & \cmark & \cmark    & \best{33.83} & \best{0.9450}  & \best{0.0204}   \\
    Ours$_\text{w/o P3D}$ & \cmark  & \xmark    &   32.51 & 0.9270 & 0.0326  \\
    Ours$_\text{w/o HAR \& P3D}$ &\xmark  & \xmark   &  29.49 & 0.8670 & 0.0426 \\
  \bottomrule
  \end{tabular}
  }
\label{tab:decoder}
\vspace{-0.5em}
\end{table}
%%%%%%%%%%%%%%%%%%%%%%%%%%%%%%%%%%%%%%%%%%%%%%%%%%
%%%%%%%%%%%%%%%%%%%%%%%%%%%%%%%%%%%%%%%%%%%%%%%%%
%%%%%%%%%%%%%%%%%%%%%%%%%%%%%%%%%%%%%%%%%%%%%%%%%

\begin{figure}[t]
    \centering
    \includegraphics[width=1\linewidth]{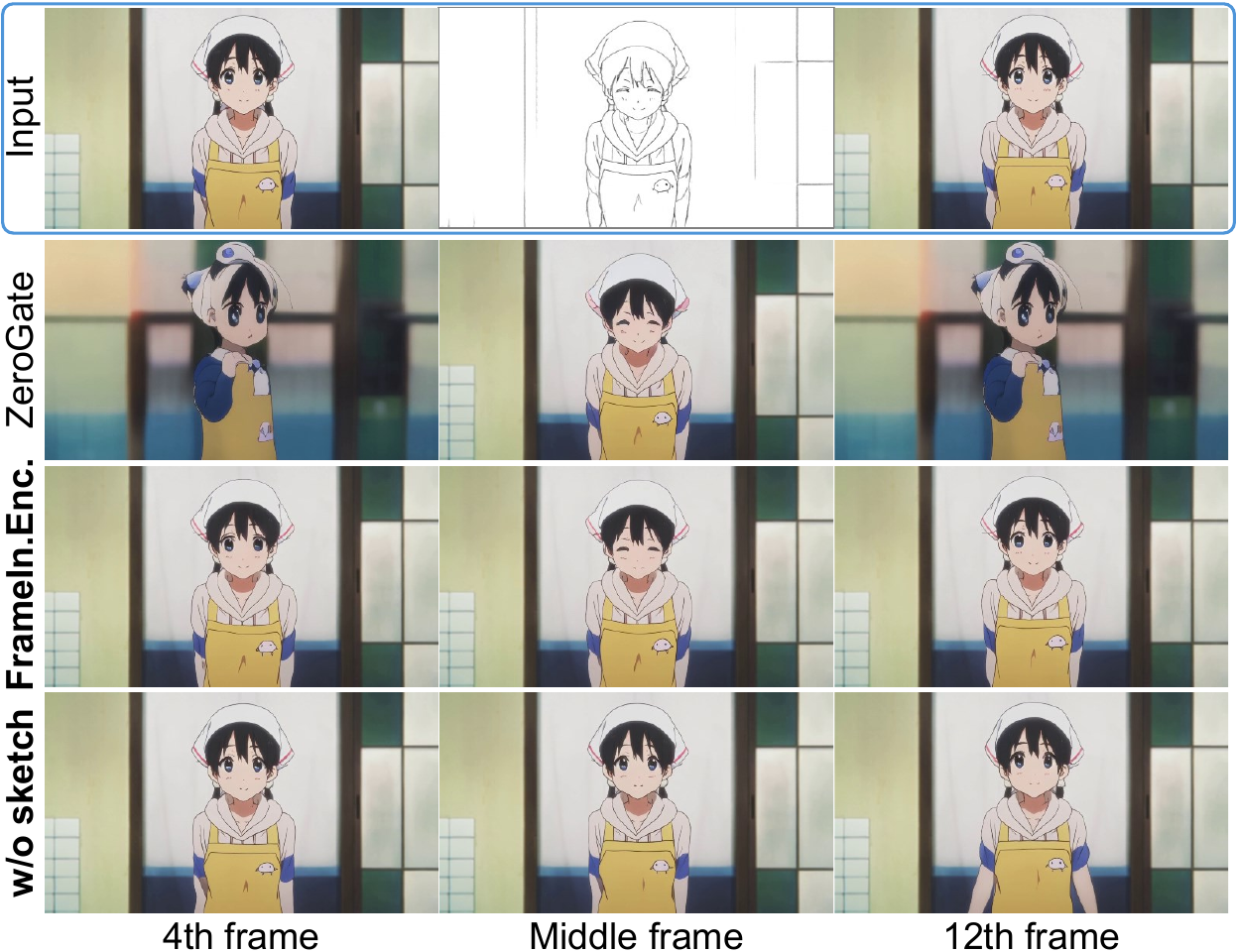}
    \vspace{-1em}
    \caption{
    Visual comparison of intermediate frames generated by different variants, with or without sparse sketch guidance (middle frame only).
    }
    \label{fig:ablation_sketch}
\vspace{-0.5em}
\end{figure}

\textbf{\textit{Sparse sketch guidance.}}
To verify the design of our frame-independent sketch encoder, we construct a variant by training $\mathcal{S}$ with full conditions $\mathbf{s}$ (\ie, per-frame sketch guidance) and enable its sparse control by zeroing out $\mathbf{F}^i_\text{inject}$ for frames without guidance during inference.
We provide only the middle-frame sketch as sparse control and compare this \textbf{ZeroGate} variant with our \textbf{FrameIn.Enc.} (frame-independent encoder), as shown in Figure~\ref{fig:ablation_sketch}. Although `ZeroGate' can generate the middle frame adhering to the sketch guidance, it struggles to produce consistent content for other unguided frames. In contrast, our `FrameIn.Enc.' not only generates middle frames with good conformity to the sketch, but also maintains temporal coherence across the generated sequence. 
We also present the generated results without sketch guidance (4th row in Figure~\ref{fig:ablation_sketch}) using the same input cartoon images.

%%%%%%%%%%%%%%%%%%%%%%%%%%%%%%%%%%%%%%%%%%%%%%%%%
%%%%%%%%%%%%%%%%%%%%%%%%%%%%%%%%%%%%%%%%%%%%%%%%%

%%%%%%%%%%%%%%%%%%%%%%%%%%%%%%%%%%%%%%%%%%%%%%%%%
% index=4,11,18 for the three sketches
%%%%%%%%%%%%%%%%%%%%%%%%%%%%%%%%%%%%%%%%%%%%%%%%%

\section{Applications}
\begin{figure*}[t]
    \centering
    \includegraphics[width=0.98\linewidth]{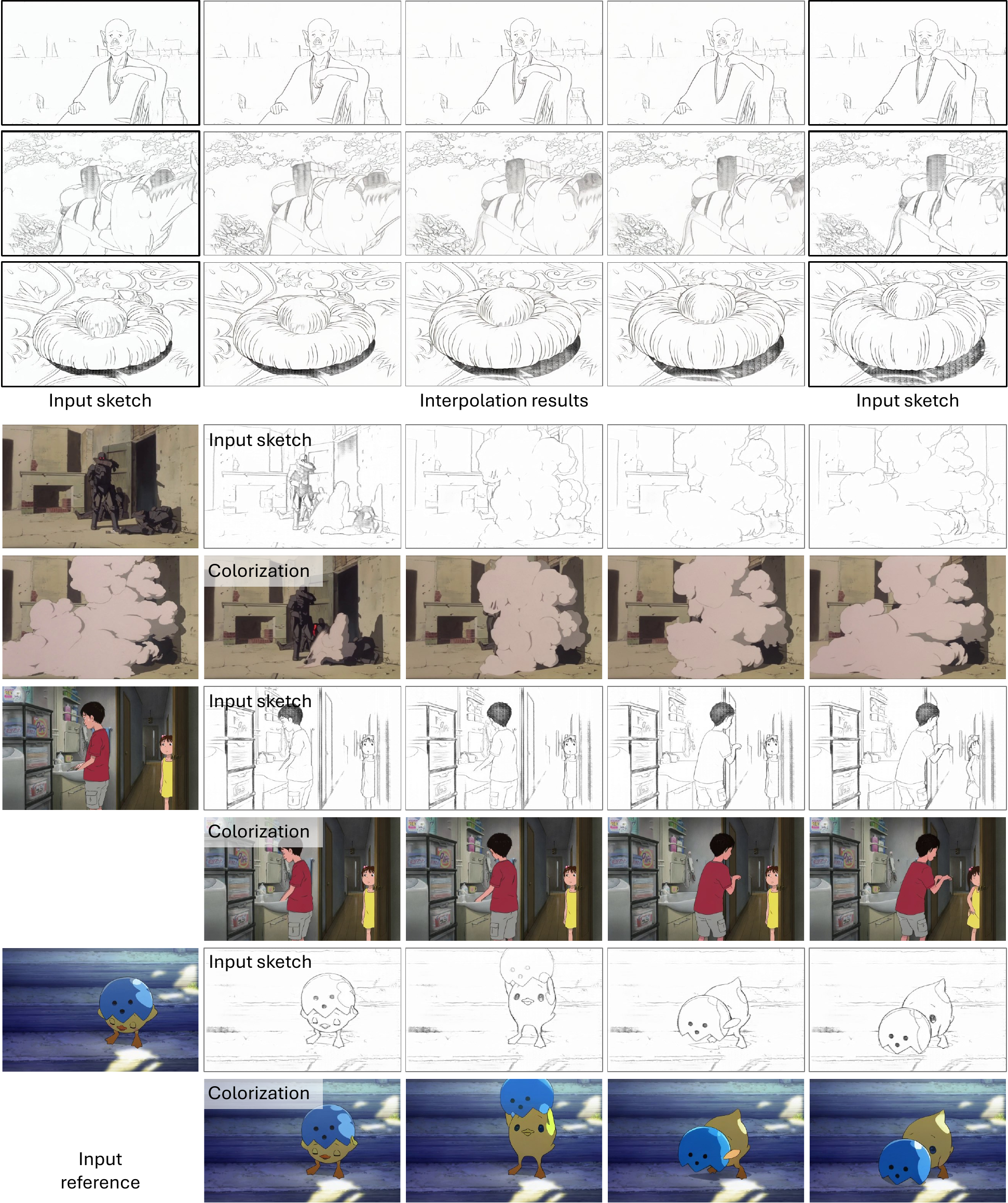}
    \caption{
    Results when our method is applied for: (upper row) cartoon sketch interpolation, and (middle and lower rows) reference-based sketch colorization.
    }
    \label{fig:application}
    % \vspace{-3mm}
\end{figure*}

\textbf{Cartoon sketch interpolation} is more than challenging due to its extremely sparse structure without color and textures. Nonetheless, our ToonCrafter can still produce decent results (Figure~\ref{fig:application} top) on such unseen input, thanks to its powerful generalization capabilities.
It can also support \textbf{reference-based sketch colorization} by providing  1 or 2 reference images and per-frame sketches. The visual results of these applications are presented in Figure~\ref{fig:application}.

\section{Conclusion}
\label{sec:conclusion}
We introduced \emph{ToonCrafter}, an innovative framework for the first attempt of generative cartoon interpolation. 
We propose the toon rectification learning to retain the live-action motion priors while overcoming the domain gap, and preserve the visual details
through the dual-reference-based 3D decoder. To allow user control over the interpolation, we design a frame-independent sketch encoder. Both quantitative and qualitative experimental results evidence the effectiveness and superiority of our method compared to existing competitors. Lastly, we showcase the versatility of our approach across various applications.

% \clearpage

%%%%%%%%%%%%%%%%%%%%%%%%%%%%%%%%%%%%%%%%%%%%%%%%%
%%%%%%%%%%%%%%%%%%%%%%%%%%%%%%%%%%%%%%%%%%%%%%%%%

%%%%%%%%%%%%%%%%%%%%%%%%%%%%%%%%%%%%%%%%%%%%%%%%%
%%%%%%%%%%%%%%%%%%%%%%%%%%%%%%%%%%%%%%%%%%%%%%%%%

%%%%%%%%% REFERENCES
{\small
\bibliographystyle{ieee_fullname}
\bibliography{egbib}
}

\end{document}